\documentclass{ifacconf}
\usepackage{amsmath}
\usepackage{siunitx}
\usepackage{amsfonts,amssymb,color}  
\usepackage{graphicx}      
\usepackage{subcaption}
\usepackage{natbib}        
\newcommand{\bzero}{\ensuremath{\mathbf{0}}}
\newcommand{\real}{\mathbb{R}}
\newcommand{\bOmega}{\ensuremath{\mathbf{\Omega}}}
\newcommand{\bSigma}{\ensuremath{\mathbf{\Sigma}}}
\newcommand{\bM}{\ensuremath{\mathbf{M}}}
\newcommand{\bA}{\ensuremath{\mathbf{A}}}
\newcommand{\bK}{\ensuremath{\mathbf{K}}}
\newcommand{\bE}{\ensuremath{\mathbf{E}}}
\newcommand{\bC}{\ensuremath{\mathbf{C}}}
\newcommand{\bQ}{\ensuremath{\mathbf{Q}}}
\newcommand{\bu}{\ensuremath{\mathbf{u}}}
\newcommand{\bU}{\ensuremath{\mathbf{U}}}
\newcommand{\bB}{\ensuremath{\mathbf{B}}}
\newcommand{\bD}{\ensuremath{\mathbf{D}}}
\newcommand{\bY}{\ensuremath{\mathbf{Y}}}
\newcommand{\bV}{\ensuremath{\mathbf{V}}}
\newcommand{\tbA}{\tilde{\bA}}
\newcommand{\tbB}{\tilde{\bB}}
\newcommand{\tbE}{\tilde{\bE}}
\newcommand{\tbq}{\tilde{\bq}}

\newcommand{\bbA}{\bar{\bA}}
\newcommand{\bbB}{\bar{\bB}}
\newcommand{\bbE}{\bar{\bE}}
\newcommand{\bbx}{\bar{\bx}}
\newcommand{\bbD}{\bar{\bD}}
\newcommand{\hbQ}{\widehat{\bQ}}
\newcommand{\hbY}{\widehat{\bY}}

\newcommand{\hbE}{\widehat{\bE}}
\newcommand{\hbK}{\widehat{\bK}}
\newcommand{\hbC}{\widehat{\bC}}
\newcommand{\hbB}{\widehat{\bB}}

\newcommand{\bq}{\ensuremath{\mathbf{q}}}
\newcommand{\hbq}{\widehat{\bq}}
\newcommand{\by}{\ensuremath{\mathbf{y}}}
\newcommand{\bx}{\ensuremath{\mathbf{x}}}
\newcommand{\bff}{\ensuremath{\mathbf{f}}}

\newcommand{\hbqd}{\dot{\hbq}}
\newcommand{\hbqdd}{\ddot{\hbq}}
\newcommand{\bqd}{\dot{\bq}}

\newcommand{\hbQd}{\dot{\hbQ}}
\newcommand{\hbQdd}{\ddot{\hbQ}}
\newcommand{\Ra}[1]{\color{black} {#1}}
\newcommand{\Rb}[1]{\color{black} {#1}}
\usepackage{tikz}
\usepackage{tikz-imagelabels}
\usepackage{pgfplots} 
\usepackage{pgfgantt}
\usepackage{pdflscape}
\pgfplotsset{compat=newest} 
\pgfplotsset{plot coordinates/math parser=false} 
\newlength\fwidth
\newlength\fheight
\begin{document}
\begin{frontmatter}

\title{Data-driven Model Reduction for Soft \\ Robots via Lagrangian Operator Inference} 


\author[First]{Harsh Sharma} 
\author[First]{Iman Adibnazari} 
\author[First]{Jacobo Cervera-Torralba}
\author[First]{ Michael T. Tolley}
\author[First]{Boris Kramer}

\address[First]{Department of Mechanical and Aerospace Engineering, University of California San Diego, La Jolla, California, USA \\
\{hasharma, iadibnaz, jcerveratorralba, mtolley, bmkramer\}@ucsd.edu.}

\begin{abstract}                
Data-driven model reduction methods provide a nonintrusive way of constructing computationally efficient surrogates of high-fidelity models for real-time control of soft robots. This work leverages the Lagrangian nature of the model equations to derive structure-preserving linear reduced-order models via Lagrangian Operator Inference and compares their performance with prominent linear model reduction techniques through an anguilliform swimming soft robot model example with $231{,}336$ degrees of freedom. The case studies demonstrate that preserving the underlying Lagrangian structure leads to learned models with higher predictive accuracy and robustness to unseen inputs.
\end{abstract}


\end{frontmatter}

\section{Introduction}
Soft robotics is an emerging field of robotics that leverages the natural compliance of soft materials to build robots for a variety of tasks such as grasping a wide range of unknown objects, achieving energy-efficient movement in complex environments, safely interacting with humans, and building haptic interfaces (\cite{rus2015design}). Soft robots are inherently infinite-dimensional dynamical systems and due to their compliance, they often exhibit strong nonlinear dynamics. Both the infinite-dimensional nature and the strong nonlinear behavior of the soft robot dynamics make their modeling and control challenging, see~\cite{della2023model} for more details about challenges in model-based control of soft robots.

The most common soft robot modeling approach is to work with reduced-physics models obtained via simplifying assumptions. By assuming that the strain is piecewise constant along the arc length, piecewise constant curvature models and their extensions have achieved significant success in soft robot modeling, see~\cite{webster2010design} for more details. In another research direction, models based on functional parametrizations have been widely used for modeling flexible link robots (\cite{de2016robots}). Recently, a variety of soft robot models based on the Cosserat rod theory have been used for describing the dynamics of tendon-actuated soft robots in~\cite{renda2018discrete,janabi2021cosserat}. Even though the reduced-physics models are well-suited for control purposes, all of these modeling approaches are problem-specific which restricts their broad applicability.

As an alternative to simplified soft robot models, simulators based on the finite element method (FEM) provide high-fidelity models of soft robots, which typically leads to high-dimensional full-order models (FOMs) with hundreds of thousands of degrees of freedom (DOF). This high-dimensional nature of the soft robot FOMs makes their real-time control computationally challenging. 

Projection-based reduced-order models (ROMs) have been proposed to derive fast low-dimensional surrogates of high-dimensional soft robot models.  In~\cite{thieffry2018control}, the authors derived projection-based linear ROMs of a nonlinear soft beam model with $5{,}442$ DOF. The authors in~\cite{goury2018fast} demonstrated real-time control of a nonlinear tentacle robot through the use of a projection-based nonlinear model reduction approach with hyperreduction. In~\cite{katzschmann2019dynamically}, they employed projection-based linear ROMs for controlling a pneumatically-actuated soft robot arm with $70{,}860$ DOF. A high-dimensional linear benchmark model for a soft robotic fishtail with $779{,}232$ DOF has been developed in~\cite{siebelts2018modeling} and been used for a comparative study of projection-based second-order model reduction methods in~\cite{saak2019comparison}. In~\cite{tonkens2021soft}, the authors proposed a piecewise affine model reduction strategy for reduced-order modeling of a cable-driven nonlinear soft robot model with $9{,}768$ DOF. However, all of these model reduction approaches are \textit{intrusive} in the sense that they require access to the high-dimensional FOM operators from the simulator. 

The governing equations for soft robot dynamics can be derived from a Lagrangian mechanics perspective, which provides a rich mechanical and geometric structure for the high-fidelity soft robot models. In this work, we advocate a \textit{nonintrusive} and \textit{structure-preserving} model reduction approach for soft robotics applications where we exploit the Lagrangian structure of these models to derive data-driven Lagrangian ROMs via Lagrangian Operator Inference. Since our main goal is to make model predictive control of soft robots computationally tractable in real time, we focus on learning linear ROMs. Towards this goal, we derive data-driven linear Lagrangian ROMs of a high-dimensional nonlinear soft robot model with $231{,}336$ DOF and evaluate the numerical performance of the proposed approach against two widely used data-driven linear model reduction techniques-- dynamic mode decomposition with control and the eigensystem realization algorithm. The nonintrusive structure-preserving model reduction approach proposed in this work is particularly relevant for soft robot applications where full access to---and the ability to manipulate---the source code and the underlying dynamical system operators is often not possible. 

This paper is structured as follows. Section~\ref{sec:background} provides a brief introduction to high-dimensional soft robot models and reviews three methods for deriving data-driven linear ROMs. Section~\ref{sec:case} describes the high-dimensional soft robot model considered for the comparative study.  Section~\ref{sec: conclusions} provides concluding remarks and future research directions.
\section{Background}
\label{sec:background}
In Section~\ref{sec: FOM}, we describe the high-dimensional soft robot models considered herein. In Section~\ref{sec:lopinf}, we provide a brief overview of the Lagrangian Operator Inference method, a nonintrusive model reduction approach for deriving linear structure-preserving ROMs from data. In Section~\ref{sec:dmdc}, we summarize the dynamic mode decomposition with control method for deriving linear data-driven ROMs in the discrete-time domain. Finally, we review the eigensystem realization algorithm for deriving data-driven linear ROMs purely from input-output data in Section~\ref{sec: era}.
\subsection{High-dimensional soft robot models}
\label{sec: FOM}
{\Rb{We focus on nonlinear soft robot models of the form}}    
\begin{equation}
        \bM \ddot{\bq}(t) + \bC \dot{\bq}(t) + \frac{\partial U}{\partial \bq}(\bq(t)) = \bB \bu(t),  
        \label{eq:fom}
    \end{equation}
where $\bq(t)$ is the state vector containing displacements of each node at spatial locations defined by the finite element mesh, $\bM \succ \bzero\in \real^{n \times n}$ is the symmetric positive-definite inertia matrix, $\bC \succ \bzero\in \real^{n \times n}$ is the symmetric positive-definite damping matrix, $U(\bq(t))$ is the scalar potential energy function, $\bB$ is the constant input matrix, and $\bu(t) \in \real^{m}$ is the time-dependent (control) input vector. {\Ra{We consider the following linear output equation}}
\begin{equation}
       \by(t)=\bE\bq(t),
\end{equation}
where $\by\in \real^{p}$ is the output vector {\Ra{containing a small set of performance variables (e.g. displacements at nodes along the soft robot's centerline)}} and $\bE \in \real^{p \times n}$ is the output matrix.

Our goal in this work is to learn a fast low-dimensional approximation of~\eqref{eq:fom} from data. To this end, given a time-dependent input vector $\bu(t)$ and initial conditions $(\bq(0),\bqd(0))$, let $\bq_1, \cdots, \bq_K$ be the solutions to~\eqref{eq:fom} at $t_1, \cdots, t_K$ computed using a soft robot simulator. Let $\by_1, \cdots, \by_K$ be the corresponding outputs at those time instances. We first build the FOM snapshot data matrix
\begin{equation}
    \bQ=[\bq_1, \cdots, \bq_K] \in \real^{n \times K}.
    \label{eq:snapshot}
\end{equation}
We also define the corresponding input and output snapshot matrices
\begin{equation}
    \bU=[\bu_1,\cdots,\bu_K] \in \real^{m \times K}, \quad \bY=[\by_1, \cdots, \by_K] \in \real^{p \times K},
      \label{eq:snapshot_input}
\end{equation}
where $\bu_k:=\bu(t_k)$ is the input vector at time $t_k$.
\subsection{Lagrangian Operator Inference method}
\label{sec:lopinf}
The Lagrangian Operator Inference (LOpInf) method (\cite{sharma2022preserving}) is a nonintrusive model reduction method that constrains the learned reduced models to have Lagrangian mechanical structure. The method first projects high-dimensional snapshot data onto a low-dimensional subspace and then fits linear reduced operators to the reduced data in a structure-preserving way.  {\Ra{Assuming the high-dimensional snapshot data admits an accurate representation in low-dimensional linear subspaces}}, we first construct a low-dimensional basis matrix  $\bV_r\in \real^{n \times r}$ from the FOM snapshot data matrix $\bQ$ in~\eqref{eq:snapshot} via singular value decomposition (SVD). {\Ra{The reduced dimension $r$ is typically chosen so that the leading $r$ singular values capture more than $95\%$ of the total energy}}. We then obtain reduced snapshot data via projections as  
    \begin{equation}
        \hbQ=\bV_r^\top\bQ=[\hbq_1, \cdots, \hbq_K] \in \real^{r \times K},
        \label{eq:reduced_data}
    \end{equation}
    where $\hbq_k:=\bV_r^\top\bq_k$ is the reduced snapshot at time $t_k$. We also build snapshot matrices of the reduced first-order and second-order time-derivative data
    \begin{equation*}
          \dot{\hbQ}=[\dot{\hbq}_1, \cdots, \dot{\hbq}_K] \in \real^{r \times K}, \quad  \ddot{\hbQ}=[\ddot{\hbq}_1, \cdots, \ddot{\hbq}_K] \in \real^{r \times K},
    \end{equation*}
    where $\dot{\hbq_k}$ and $\ddot{\hbq}_k$ are obtained from the reduced snapshot data $\hbQ$ in~\eqref{eq:reduced_data} using a finite difference scheme, e.g., an eighth-order central finite difference scheme. 
    
We postulate  a reduced Lagrangian with the reduced nonconservative forcing 
 \begin{equation}
\widehat{L}(\hbq,\dot{\hbq})=\frac{1}{2}\hbqd^\top\hbqd - \frac{1}{2}\hbq^\top\hbK\hbq, \quad \widehat{\bff}(\hbqd,t)=-\hbC\hbqd - \hbB\bu(t),
\label{eq:lhat}
 \end{equation}
with the reduced damping matrix $\hbC \in \real^{r \times r}$, the reduced stiffness matrix $\hbK \in \real^{r \times r}$, and the reduced input matrix $\hbB \in \real^{r \times m}$. Substituting expressions for the reduced Lagrangian $\widehat{L}(\hbq,\dot{\hbq})$ and the reduced nonconservative forcing $ \widehat{\bff}(\hbqd,t)$ from~\eqref{eq:lhat} into the forced Euler-Lagrange equations
\begin{equation}
\frac{\partial \widehat{L}(\hbq,\dot{\hbq})}{\partial \hbq} - \frac{\text {d}}{\text {d}t}\left( \frac{\partial \widehat{L}(\hbq,\dot{\hbq})}{\partial \dot{\hbq}} \right) + \widehat{\bff}(\hbqd,t) =\bzero,
\end{equation}
yields the governing equations for the Lagrangian ROM
\begin{equation}
      \hbqdd(t)+ \hbC \hbqd (t) + \hbK\hbq(t)=\hbB \bu(t).
        \label{eq:L_rom}
 \end{equation}
The corresponding output equation for the Lagrangian ROM is $\by(t)=\hbE \hbq (t)$ where $\hbE \in \real^{p \times r}$ is the reduced output matrix.
Given the reduced snapshot data $\hbQ$ and the reduced time-derivative data $\dot{\hbQ}$ and $\ddot{\hbQ}$, we infer the Lagrangian ROM operators in~\eqref{eq:L_rom} by solving
\begin{equation}
\min _{\substack{\hbK=\hbK^\top \succ 0, \hbC=\hbC^\top \succ 0, \hbB}}
\lVert \hbQdd + \hbC\hbQd + \hbK\hbQ- \hbB \bU \rVert_{F}.
\label{eq:lopinf}
\end{equation}
The reduced output operator $\hbE$ is inferred by solving the least-squares problem 
\begin{equation}
    \min _{\hbE}
\lVert \bY - \hbE\hbQ \rVert_{F}.
\label{eq:lopinf_output}
\end{equation}
{\Rb{The constrained optimization problem in~\eqref{eq:lopinf} is solved using the CVX optimization package (\cite{cvx})}} where the symmetric positive-definite constraints on $\hbK$ and $\hbC$ in~\eqref{eq:lopinf} ensure that the linear ROM~\eqref{eq:L_rom} preserves the underlying Lagrangian structure. In parallel to LOpInf, a second-order Operator Inference oriented approach has been developed in~\cite{filanova2023operator} where the authors propose solving an unconstrained optimization problem with a post-processing procedure to obtain reduced mass, stiffness, and damping matrices. In the numerical studies, we will compare LOpInf with two other prominent linear data-driven model reduction/system ID approaches, which we outline next.
\subsection{Dynamic mode decomposition with control}
\label{sec:dmdc}
Dynamic mode decomposition with control (DMDc)(\cite{proctor2016dynamic}) is a nonintrusive method to learn linear ROMs with external forcing of high-dimensional complex systems purely from data. The first step is to build the snapshot data matrix $\bQ=[\bq_1, \cdots, \bq_{K-1}] \in \real^{n \times K-1}$, and the time-shifted snapshot data matrix $ \bQ'=[\bq_2, \cdots, \bq_{K}] \in \real^{n \times K-1}$. The corresponding input snapshot matrix is $\bU=[\bu_1,\cdots, \bu_{K-1}]\in \real^{m \times K-1}$. The data marices $\bQ$ and $\bU$ are stacked to construct $\bOmega:=[ \bQ^\top,  \bU^\top ]^\top \in \real^{n+m \times K-1}$. Given data matrices $\bOmega$ and $\bQ'$, we compute the SVD of $\bOmega$ with truncation value $p$ and compute the SVD of $\bQ'$ with truncation value~$r$: 
\begin{equation}
    \bOmega\approx \begin{bmatrix} \tilde{\bU}_1 \\ \tilde{\bU}_2  \end{bmatrix}\tilde{\bSigma}\tilde{\bV}^*,\quad  \bQ'\approx \widehat{\bU}\widehat{\bSigma}\widehat{\bV}^*,
\end{equation}
with $\tilde{\bU}_1 \in \real^{n \times p}$, $\tilde{\bU}_2 \in \real^{m \times p}$, $\tilde{\bSigma} \in \real^{p \times p}$, $\tilde{\bV}^*\in \real^{p \times K-1}$, $ \widehat{\bU} \in \real^{n \times r}$, $\widehat{\bSigma} \in \real^{r \times r}$, and $\widehat{\bV}^* \in \real^{r \times K-1}$. Using the FOM state approximation $\bq\approx\widehat{\bU}\tbq$, the resulting linear DMDc ROM is
    \begin{equation}
        \tbq_{k+1}=\tbA\tbq_k + \tbB \bu_k, \qquad \by_{k+1}=\tbE \tbq_{k+1},
    \end{equation}
with the reduced state matrix $\tbA:=\widehat{\bU}^*\bQ'\tilde{\bV}\tilde{\bSigma}^{-1}\tilde{\bU}_1^*\widehat{\bU} \in \real^{r \times r}$, the reduced input matrix $\tbB:=\widehat{\bU}^*\bQ'\tilde{\bV}\tilde{\bSigma}^{-1}\tilde{\bU}_2^* \in \real^{m \times r}$, and the reduced output matrix $\tbE:=\bE_{\text{approx}}\widehat{\bU}$ where $\bE_{\text{approx}}\in \real^{p\times n}$ is computed by solving 
\begin{equation}
    \min _{\bE_{\text{approx}}}
\lVert \bY - \bE_{\text{approx}}\bQ \rVert_{F}.
\label{eq:dmdc_output}
\end{equation}
\subsection{Eigensystem realization algorithm}
\label{sec: era}
The Eigensystem realization algorithm (ERA)(\cite{kung1978new,juang1985eigensystem}) is a system identification method for identifying low-dimensional linear input-output models from impulse response data. For arbitrary input-output data, the ERA approach is used in combination with the observer Kalman filter identification (OKID) algorithm (\cite{juang1993identification}). Unlike LOpInf and DMDc, the ERA/OKID approach only requires the input-output data and does not require the FOM snapshot data matrix. Given snapshot data matrices $\bU$ and $\bY$ from~\eqref{eq:snapshot_input}, the ERA/OKID procedure yields a linear ROM of the form
\begin{equation}
    \bbx_{k+1}=\bbA\bbx_k + \bbB \bu_k, \qquad \by_{k+1}=\bbE \bbx_{k+1} + \bbD \bu_{k+1},
\end{equation}
with reduced state matrix $\bbA \in \real^{r \times r}$, reduced input matrix $\bbB \in \real^{r \times m}$, reduced output matrix $\bbE \in \real^{p \times r}$, and feedforward matrix $\bbD \in \real^{p \times m}$. 

\section{Soft robot case study}
\label{sec:case}
Anguilliform swimming is an energy-efficient mode of underwater locomotion utilized by elongated fish like eels and oarfish. The morphological simplicity of their bodies along with their maneuverability and energy-efficient characteristics makes anguilliform swimming soft robots ideal for long-duration tasks in complex, underwater environments.

We demonstrate LOpInf for a high-dimensional swimming soft robot model, described in Section~\ref{sec:fom}, and then compare its numerical performance with DMDc and ERA/OKID. In Section~\ref{sec: results_single}, we present a time extrapolation study for the case where the data-driven ROMs are learned using data from a single simulation. In Section~\ref{sec: results_multiple}, we study the data-driven ROMs' robustness to unseen inputs for the case where the training dataset is built from multiple simulations.
%
\begin{figure}[h]
    \centering
    \begin{subfigure}{0.45\linewidth}
 \includegraphics[width=\linewidth,height=3cm]{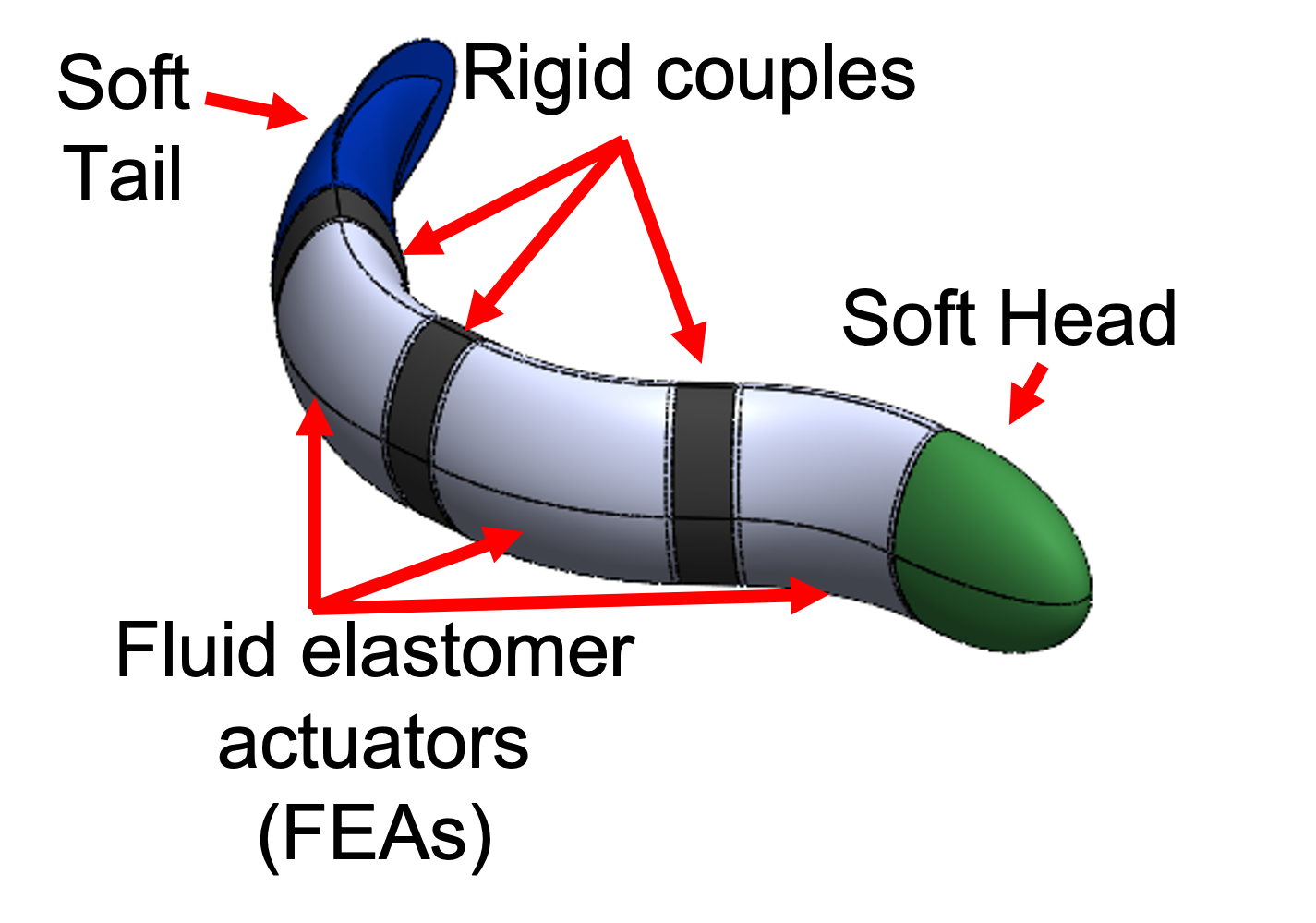}
        \caption{Schematic}
        \label{fig:robot}
    \end{subfigure}  
     \begin{subfigure}{0.45\linewidth}
 \includegraphics[width=\linewidth,height=3cm]{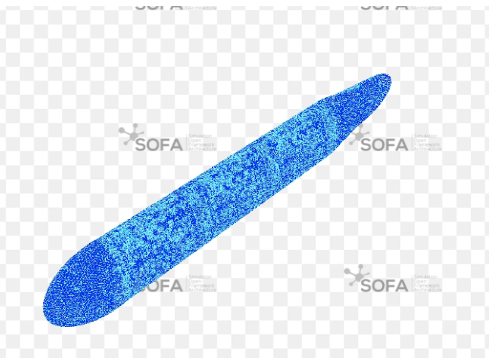}
        \caption{Three-dimensional model}
        \label{fig:sofa}
    \end{subfigure} 
    \caption{Anguilliform swimming soft robot model. }
    \label{fig:fom}
\end{figure}
\subsection{Anguilliform swimming soft robot model}
\label{sec:fom}

The FOM considered here is based on the soft robot model created in the Simulation Open Framework Architecture (SOFA)(\cite{faure2012sofa}) and considers the soft body dynamics in isolation. The simulated robot in Figure~\ref{fig:fom} is composed of soft, unactuated segments for the head and tail, three soft body segments with two antagonistic fluid elastic actuators in each segment, and rigid couples that connect each of the soft segments. The soft robot dynamics are modeled using linear elasticity constitutive laws, making the geometric nonlinearities the only source of nonlinearities in the FOM. The three-dimensional model of the soft robot is meshed with \textsf{Gmsh} (\cite{geuzaine2009gmsh}), an open-source meshing software, resulting in a FOM with $n = 231{,}336$ DOF. For this study, the time-dependent input vector $\bu(t)$ of dimension $m=6$ corresponds to the pressure inputs to each of the six fluid chambers in the fluid elastomer actuators (see Figure~\ref{fig:robot}), and the output vector $\by$ of dimension $p=40$ corresponds to the $x-$ and $z-$coordinates of points along the soft robot's centerline.
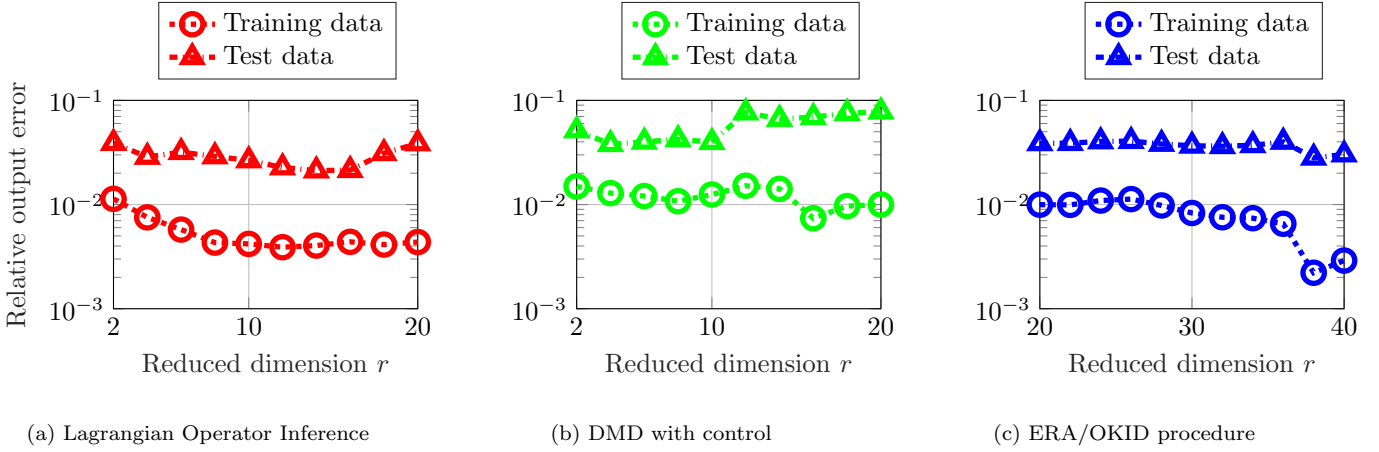
\begin{figure*}[t]
    \begin{subfigure}{0.3\linewidth}
         \setlength\fheight{4.25 cm}
         \setlength\fwidth{\textwidth}
%
%
\begin{tikzpicture}

\begin{axis}[%
width=0.942\fheight,
height=0.65\fheight,
at={(0\fheight,0\fheight)},
scale only axis,
xmin=2,
xtick={2,10,20},
xmax=20,
xlabel style={font=\color{white!15!black}},
xlabel={Reduced dimension $r$},
ymode=log,
ymin=0.001,
ymax=0.1,
yminorticks=true,
ylabel style={font=\color{white!15!black}},
ylabel={Relative output error},
axis background/.style={fill=white},
xmajorgrids,
ymajorgrids,
legend style={at={(0.147,1.1)}, anchor=south west, legend cell align=left, align=left, draw=white!15!black}
]
\addplot [color=red, dotted, line width=2.0pt, mark size=4.0pt, mark=o, mark options={solid, red}]
  table[row sep=crcr]{%
2	0.0113356469284373\\
4	0.00753244988189542\\
6	0.00575081017387661\\
8	0.00431240643811969\\
10	0.00419025825141938\\
12	0.00389619544848317\\
14	0.00404029974789915\\
16	0.00438441864560886\\
18	0.00412886650721988\\
20	0.0043640087158089\\
22	0.0047626803221713\\
};
\addlegendentry{Training data};
\addplot [color=red, dashdotted, line width=2.0pt, mark size=4pt, mark=triangle, mark options={solid, red}]
  table[row sep=crcr]{%
2	0.0388928145401896\\
4	0.0286408413198126\\
6	0.0318352921840387\\
8	0.0289693865961626\\
10	0.0265172679672574\\
12	0.0226744205369132\\
14	0.0211889371066334\\
16	0.0213819591957921\\
18	0.0310963534969727\\
20	0.0386063124246519\\
22	0.0382196868892179\\
};
\addlegendentry{Test data};
\end{axis}

\begin{axis}[%
width=1.516\fheight,
height=1.137\fheight,
at={(-0.381\fheight,-0.302\fheight)},
scale only axis,
xmin=0,
xmax=1,
ymin=0,
ymax=1,
axis line style={draw=none},
ticks=none,
axis x line*=bottom,
axis y line*=left
]
\end{axis}
\end{tikzpicture}%
        \caption{Lagrangian Operator Inference}
        \label{fig:output_error_lopinf}
    \end{subfigure}  \hspace{0.4cm}
    \begin{subfigure}{0.3\linewidth}
         \setlength\fheight{4.25 cm}
         \setlength\fwidth{\textwidth}
%
%
\begin{tikzpicture}

\begin{axis}[%
width=0.942\fheight,
height=0.65\fheight,
at={(0\fheight,0\fheight)},
scale only axis,
xmin=2,
xmax=20,
xtick={2,10,20},
xlabel style={font=\color{white!15!black}},
xlabel={Reduced dimension $r$},
ymode=log,
ymin=0.001,
ymax=0.1,
yminorticks=true,
ylabel style={font=\color{white!15!black}},
axis background/.style={fill=white},
xmajorgrids,
ymajorgrids,
legend style={at={(0.147,1.1)}, anchor=south west, legend cell align=left, align=left, draw=white!15!black}
]
\addplot [color=green, dotted, line width=2.0pt, mark size=4.0pt, mark=o, mark options={solid, green}]
  table[row sep=crcr]{%
2	0.0149199026798877\\
4	0.0128717891367139\\
6	0.0120697816975013\\
8	0.0107635446144707\\
10	0.0124834116509983\\
12	0.0151504243869341\\
14	0.0141351593551639\\
16	0.00739015350538333\\
18	0.00966593735864532\\
20	0.0100063190994885\\
22	0.00662633415641219\\
};
\addlegendentry{Training data};
\addplot [color=green, dashdotted, line width=2.0pt, mark size=4pt, mark=triangle, mark options={solid, green}]
  table[row sep=crcr]{%
2	0.0514267389208633\\
4	0.0379657065518211\\
6	0.0400225591610161\\
8	0.042281743971572\\
10	0.0397181244478321\\
12	0.0763548962463701\\
14	0.066276785232915\\
16	0.0690608726336796\\
18	0.0754524992645029\\
20	0.0779591820292529\\
22	0.0696969477564398\\
};
\addlegendentry{Test data};
\end{axis}

\begin{axis}[%
width=1.516\fheight,
height=1.137\fheight,
at={(-0.381\fheight,-0.302\fheight)},
scale only axis,
xmin=0,
xmax=1,
ymin=0,
ymax=1,
axis line style={draw=none},
ticks=none,
axis x line*=bottom,
axis y line*=left
]
\end{axis}
\end{tikzpicture}%
        \caption{DMD with control}
        \label{fig:output_error_dmdc}
    \end{subfigure}  \hspace{0.4cm}
        \begin{subfigure}{0.3\linewidth}
         \setlength\fheight{4.25 cm}
         \setlength\fwidth{\textwidth}
%
%
\begin{tikzpicture}

\begin{axis}[%
width=0.942\fheight,
height=0.65\fheight,
at={(0\fheight,0\fheight)},
scale only axis,
xmin=20,
xmax=40,
xtick={20,30,40},
xlabel style={font=\color{white!15!black}},
xlabel={Reduced dimension $r$},
ymode=log,
ymin=0.001,
ymax=0.1,
yminorticks=true,
ylabel style={font=\color{white!15!black}},
axis background/.style={fill=white},
xmajorgrids,
ymajorgrids,
legend style={at={(0.147,1.1)}, anchor=south west, legend cell align=left, align=left, draw=white!15!black}
]
\addplot [color=blue, dotted, line width=2.0pt, mark size=4.0pt, mark=o, mark options={solid, blue}]
  table[row sep=crcr]{%
20	0.00997168951846531\\
22	0.00992750349780855\\
24	0.0109194418289673\\
26	0.0112799905125285\\
28	0.00980024622418861\\
30	0.00829932328373974\\
32	0.00755429156910461\\
34	0.00739438266929393\\
36	0.0065368384425641\\
38	0.00221414182074743\\
40	0.00289948987026676\\
42	0.00193098971773344\\
};
\addlegendentry{Training data};
\addplot [color=blue, dashdotted, line width=2.0pt, mark size=4pt, mark=triangle, mark options={solid, blue}]
  table[row sep=crcr]{%
20	0.0387047995268011\\
22	0.0388632759370651\\
24	0.04033420065165\\
26	0.040623064963835\\
28	0.0381735532812384\\
30	0.0366522216625328\\
32	0.036295707347972\\
34	0.0370740415298702\\
36	0.0399361299746928\\
38	0.0281305637017176\\
40	0.0302174515706687\\
42	0.027823107991933\\
};
\addlegendentry{Test data};
\end{axis}
\begin{axis}[%
width=1.516\fheight,
height=1.137\fheight,
at={(-0.381\fheight,-0.302\fheight)},
scale only axis,
xmin=0,
xmax=1,
ymin=0,
ymax=1,
axis line style={draw=none},
ticks=none,
axis x line*=bottom,
axis y line*=left
]
\end{axis}
\end{tikzpicture}%
        \caption{ERA/OKID procedure}
        \label{fig:output_error_era}
    \end{subfigure}
    \caption{Time extrapolation study (training with data from \textsf{Ep 1}). LOpInf ROMs and ERA/OKID ROMs generally achieve lower output error than the DMDc ROMs for both training and testing regimes.}
    \label{fig:output_error}
\end{figure*}
 

The numerical studies in this work consider $N=6$ simulation episodes, denoted \textsf{Ep 1} -- \textsf{Ep 6}, where the soft body is stationary at initial time $t=0$~\si{\s}, i.e., $\bq(0)=\bq_{\text{ref}}, \dot{\bq}(0)=\bzero$, and where random control sequences are applied from initial time $t=0$~\si{\s} to final time $T=2$~\si{\s}. Each simulation episode yields a high-dimensional dataset of $K=2{,}000$ snapshots which is obtained by numerically integrating the nonlinear FOM until $T=2$~\si{\s} using the in-built \texttt{EulerImplicitSolver} in SOFA with fixed time step $\Delta t=0.001$~\si{\s}. 
\subsection{Time extrapolation when trained on a single episode}
\label{sec: results_single}
We consider a setup with a single training episode and build a training dataset of $K=1{,}500$ snapshots from \textsf{Ep 1} by collecting the simulated data until time $t=1.5$~\si{\s}. For test data, we consider the FOM solution snapshots from $t=1.5$~\si{\s} to $t=2$~\si{\s} in \textsf{Ep 1}. The output error plots reported in this section compute the
\begin{equation}
    \text{Relative output error}=\lVert \bY - \hbY  \rVert_F / \lVert \bY \rVert_F,
    \label{eq:err}
\end{equation}
where $\hbY$ is the output approximation obtained from a ROM, which can be either LOpInf ROM, DMDc ROM, or ERA/OKID ROM.

In Figure~\ref{fig:output_error}, we compare the relative output error~\eqref{eq:err} for both the training data and the test data for different values of the reduced dimension. Since the three approaches considered in this comparison achieve the highest accuracy for different reduced dimension values, we show plots only for the range of values where each method achieves its highest accuracy, i.e., reduced dimension $r=2$ to $r=20$ for LOpInf and DMDc and reduced dimension $r=20$ to $r=40$ for ERA/OKID. Compared to the DMDc ROMs, we observe that the LOpInf ROMs and the ERA/OKID ROMs achieve lower relative output error in both training and test data regimes. The comparison in Figure~\ref{fig:output_error} shows that the ERA/OKID ROM of dimension $r=38$ achieves the highest accuracy in the training data regime whereas the LOpInf ROM of dimension $r=16$ achieves the highest accuracy in the test data regime. The plots in Figure~\ref{fig:output_error_dmdc} show that the DMDc approach produces the least accurate ROMs in both training and test data regimes.

\begin{figure}
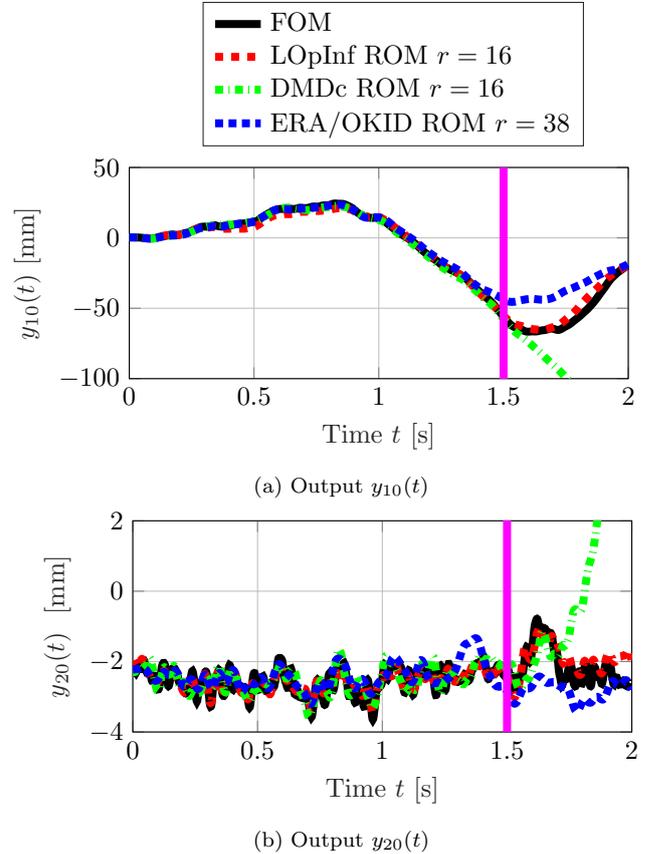

    \centering
    \vspace{0.5cm}
    \begin{subfigure}{\linewidth}
           \setlength\fheight{3.5 cm}
           \setlength\fwidth{\textwidth}
           \input{figures/y10.tex}
        \caption{Output $y_{10}(t)$}
        \label{fig:extrapolation_10}
    \end{subfigure} \\
   
    \begin{subfigure}{\linewidth}
    \centering
         \setlength\fheight{3.5 cm}
           \setlength\fwidth{\textwidth}
           \input{figures/y20.tex}
        \caption{Output $y_{20}(t)$}
        \label{fig:extrapolation_20}
    \end{subfigure}
    \caption{ Time extrapolation study (training with data from \textsf{Ep 1}). All the ROMs capture both $y_{10}(t)$ and $y_{20}(t)$ accurately in the training data regime. In the test data regime, both the LOpInf ROM and the ERA/OKID ROM provide accurate predictions whereas the DMDc ROM fails to generalize outside the training regime. The solid magenta line indicates the end of the training time interval.}
    \label{fig:extrapolation}
\end{figure}

In Figure~\ref{fig:extrapolation}, we compare the ROM solutions with the FOM solution for two representative outputs, $y_{10}$ and $y_{20}$, which are evolving at different scales. For each method, we pick the ROM with the highest accuracy in the training data regime, i.e., $r=16$ for LOpInf, $r=16$ for DMDc, and $r=38$ for ERA/OKID. We observe that all three ROMs accurately approximate the representative outputs in the training data regime. In the test data regime, we observe that the DMDc ROM fails to provide accurate predictions whereas both the LOpInf ROM and the ERA/OKID ROM track the FOM solution accurately with the LOpInf ROM still performing better. 

\subsection{Robustness to unseen control inputs}
\label{sec: results_multiple}
In this study, we consider a setup with multiple training episodes, where we build a training dataset of $K=4{,}000$ snapshots by collecting the simulated data until time $t=2$~\si{\s} from \textsf{Ep 1} and \textsf{Ep 2}. For test data, we consider the FOM solution snapshots from $t=0$~\si{\s} to $t=2$~\si{\s} for \textsf{Ep 3}, \textsf{Ep 4}, \textsf{Ep 5}, and \textsf{Ep 6}. Unlike the LOpInf method and the DMDc method, the ERA/OKID procedure is only applicable to datasets consisting of a single training episode. To ensure a fair comparison between the three data-driven approaches, we first derive the best-performing ERA/OKID ROMs from \textsf{Ep 1} and \textsf{Ep 2} separately. For the four test episodes considered in this generalization study, we observe that the ERA/OKID ROM based on \textsf{Ep 2} performs better, and hence, we compare LOpInf and DMDc ROMs based on \textsf{Ep~1} and \textsf{Ep 2} with the ERA/OKID ROM based on \textsf{Ep 2}. Similarly to the time extrapolation study in Section~\ref{sec: results_single}, we pick the ROM with the highest accuracy in the training data regime (not shown here) for each method, i.e., $r=10$ for LOpInf, $r=12$ for DMDc, and $r=28$ for ERA/OKID.

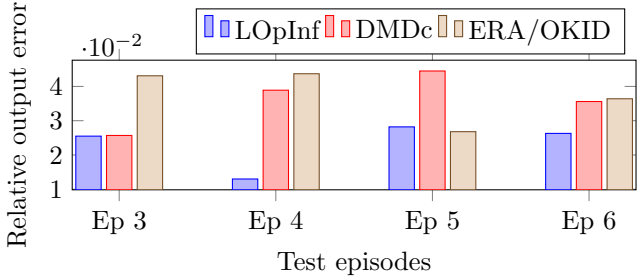
\begin{figure}
\begin{tikzpicture}  
\begin{axis}  
[      height=3.3cm,
    width=9cm,
    ybar, 
    bar width=9.5 pt,  
    legend style={at={(0.6,1.4)}, 
      anchor=north,legend columns=-2},     
    ylabel={Relative output error}, 
     xlabel={Test episodes},
    symbolic x coords={Ep 3, Ep 4, Ep 5, Ep 6},  
    xtick=data,
    ]  

\addplot coordinates { (Ep 3, 0.0255)(Ep 4,   0.0130 ) (Ep 5,0.0282) (Ep 6, 0.0263)};  

\addplot coordinates { (Ep 3, 0.0257)(Ep 4,   0.0389) (Ep 5,0.0445 ) (Ep 6, 0.0356)};

\addplot coordinates { (Ep 3, 0.0431)(Ep 4,   0.0437) (Ep 5,0.0268) (Ep 6, 0.0364)};

\legend{LOpInf, DMDc, ERA/OKID}  
  
\end{axis}  

\end{tikzpicture}  
    \caption{Robustness to unseen inputs (relative output error comparison). The LOpInf ROM achieves lower output error than both the DMDc ROM and the ERA/OKID ROM for all test episodes.}
    \label{fig:bar_plot}
\end{figure}

To get a better picture of the accuracy in the testing datasets, Figure~\ref{fig:bar_plot} provides a quantitative comparison between the three data-driven model reduction approaches for all four testing episodes. Similarly to the time extrapolation study, we observe that the structure-preserving LOpInf approach performs better than DMDc and ERA/OKID outside the training data regime.

\begin{figure}[h]
    \centering
    \begin{subfigure}{\linewidth}
         \setlength\fheight{6 cm}
         \setlength\fwidth{\textwidth}
%
%
\definecolor{mycolor1}{rgb}{1.00000,0.00000,1.00000}%
\begin{tikzpicture}

\begin{axis}[%
width=0.976\fheight,
height=0.49\fheight,
at={(0\fheight,0\fheight)},
scale only axis,
xmin=0,
xmax=1200,
xlabel style={font=\color{white!15!black}},
xlabel={$x-$coordinate [mm]},
ymin=-120,
ymax=40,
ylabel style={font=\color{white!15!black}},
ylabel={$z-$coordinate [mm]},
axis background/.style={fill=white},
xmajorgrids,
ymajorgrids,
legend style={at={(0.1,1.1)}, anchor=south west, legend cell align=left, align=left, draw=white!15!black}
]
\addplot [color=black, line width=2.0pt, mark size=3.0pt, mark=triangle, mark options={solid, black}]
  table[row sep=crcr]{%
1103.61989484187	-88.7423555230919\\
1048.96049497107	-76.7266627176095\\
991.262621764828	-64.5580082649694\\
934.85989462926	-53.2828057422764\\
876.843707542436	-40.8411156151533\\
818.898456021249	-28.8002152625677\\
760.205158766235	-15.3586427857251\\
702.195893966874	-6.52555586193898\\
645.434262223656	-2.65888142923268\\
587.701029539121	-2.02509024125993\\
527.992738760262	-1.46955874557898\\
468.682317250314	-4.27546920275336\\
408.021412979711	-2.39214696429235\\
350.42047075322	-1.81843102138305\\
289.140703886394	-3.49265072575145\\
233.948782789438	-1.85892801310797\\
176.747201586781	-3.13280681708466\\
118.475504191791	1.83662878881591\\
58.7486347857671	5.26096969917489\\
3.70743635619715	8.51774531729234\\
};
\addlegendentry{FOM}

\addplot [color=red, dotted, line width=2.0pt, mark size=3.0pt, mark=o, mark options={solid, red}]
  table[row sep=crcr]{%
1140.06121830088	-113.648599295572\\
1076.1583550373	-95.9204826666107\\
1012.62025149297	-76.1605863534719\\
951.106694910616	-59.6092178524066\\
888.012139311231	-43.2241704310291\\
824.833821129493	-28.1104250768244\\
758.790307480956	-14.3072259869361\\
700.35738857972	-5.10651876919201\\
644.250485711333	-1.62169837792794\\
587.338084027777	-2.15413414024556\\
526.476331110232	-1.46643294250271\\
469.067852570912	-5.24082668845699\\
409.415011415927	-2.55759605784169\\
351.98207404979	-0.200397002818363\\
291.713158729504	-2.16250849335188\\
234.602453603027	-0.70284251508906\\
176.177946952828	-2.3583439470342\\
118.335569753665	1.57686689062962\\
58.5759217600944	4.31367560710896\\
3.48977733452443	6.93242560254566\\
};
\addlegendentry{LOpInf ROM $r=10$}

\addplot [color=green, dashdotted, line width=2.0pt, mark size=3.0pt, mark=asterisk, mark options={solid, green}]
  table[row sep=crcr]{%
1112.11128967996	-69.6760502596753\\
1054.25860175512	-66.3880423301318\\
996.236601014113	-49.8104741720031\\
939.329148784371	-39.9797877097269\\
880.257025907988	-30.5673323806818\\
821.207102410875	-21.6309757991426\\
761.085388887341	-10.5539566874934\\
701.797854306551	-3.15854883536485\\
644.870808550636	-1.01838211068775\\
587.345871903175	-1.85763660611451\\
524.680369855467	2.27313032991037\\
465.75603083573	1.24249816939414\\
405.603225345777	5.42792449766125\\
348.330666626249	8.12869663481274\\
288.306977993399	3.6459954564782\\
232.895030139415	2.30791698096823\\
175.711281374481	-1.76989414613786\\
118.278182093337	2.7595685381666\\
58.4235777547772	7.03844121636394\\
3.25103423947917	11.0318233424694\\
};
\addlegendentry{DMDc ROM $r=12$}


\addplot [color=blue, dashed, line width=2.0pt, mark size=3.0pt, mark=+, mark options={solid, blue}]
  table[row sep=crcr]{%
1114.63280418185	-21.5548716398455\\
1059.09395637306	-17.0496140777918\\
999.658601093059	-13.1824944829254\\
941.87937661225	-10.9913549979858\\
882.131057941543	-7.46715933743212\\
822.566753235409	-4.87282236607166\\
763.023595358816	-0.727826703779101\\
703.796272067566	-0.114910462172929\\
646.201716121419	-1.20224222826459\\
587.822804976862	-2.31175814757398\\
529.416839900286	-1.29309297449095\\
470.324595245095	-4.39267154883419\\
410.236253222155	-2.08981673611402\\
352.919630769926	-1.02050031915087\\
293.97031575353	-2.83338092090617\\
235.619387778128	-1.09449664007798\\
176.219566882598	-2.52901675904036\\
118.16590726972	1.30258651033455\\
58.2523749873233	3.59069957116981\\
3.01373382628822	5.80305242714849\\
};
\addlegendentry{ERA/OKID ROM $r=28$}

\end{axis}

\begin{axis}[%
width=1.259\fheight,
height=0.743\fheight,
at={(-0.164\fheight,-0.097\fheight)},
scale only axis,
xmin=0,
xmax=1,
ymin=0,
ymax=1,
axis line style={draw=none},
ticks=none,
axis x line*=bottom,
axis y line*=left
]
\end{axis}
\end{tikzpicture}%
        \caption{$t=0.5$~\si{s}}
        \label{fig:center_test_05}
    \end{subfigure}  \\ [-4ex]
    \begin{subfigure}{\linewidth}
         \setlength\fheight{6 cm}
         \setlength\fwidth{\textwidth}
%
%
\definecolor{mycolor1}{rgb}{1.00000,0.00000,1.00000}%
\begin{tikzpicture}

\begin{axis}[%
width=0.976\fheight,
height=0.49\fheight,
at={(0\fheight,0\fheight)},
scale only axis,
xmin=0,
xmax=1200,
xlabel style={font=\color{white!15!black}},
xlabel={$x-$coordinate [mm]},
ymin=-120,
ymax=40,
ylabel style={font=\color{white!15!black}},
ylabel={$z-$coordinate [mm]},
axis background/.style={fill=white},
xmajorgrids,
ymajorgrids,
legend style={at={(0.147,0.361)}, anchor=south west, legend cell align=left, align=left, draw=white!15!black}
]
\addplot [color=black, line width=2.0pt, mark size=3.0pt, mark=triangle, mark options={solid, black}]
  table[row sep=crcr]{%
1098.09639022914	-104.448639876436\\
1044.69351555203	-88.9216936606053\\
988.124000761663	-74.069748538742\\
932.757233229592	-59.6098370543416\\
875.172276690499	-45.949067446802\\
817.659556835662	-32.4564009817425\\
760.395898223964	-18.77086255104\\
702.352015302308	-9.53544299200576\\
645.703200492397	-4.52795540135298\\
587.988054794964	-2.06573525232966\\
527.224972957357	1.18563749643567\\
469.797521305861	2.27104174115379\\
410.752348328509	3.35630005653002\\
353.87691272802	-0.95174014842928\\
292.697680186392	-0.707662225888271\\
235.041828936141	0.636371105968692\\
176.28325658947	-2.479974874562\\
118.328657214636	-3.11577031292177\\
58.7407017069688	-8.03699497981029\\
3.77383416117232	-12.3677056881386\\
};

\addplot [color=red, dotted, line width=2.0pt, mark size=3.0pt, mark=o, mark options={solid, red}]
  table[row sep=crcr]{%
1145.15888085092	-99.7023348446612\\
1077.90799216629	-78.903885516562\\
1012.91019361361	-62.8482347139395\\
950.795982364781	-48.2087780498314\\
887.315611989522	-34.0782511928094\\
823.801525576445	-21.2920095492793\\
757.677973151527	-12.3426673799513\\
699.744332292488	-4.47897084976807\\
644.222744355346	-1.7070039498924\\
587.665331300713	-1.9084515741522\\
527.127537099553	1.68584927488723\\
470.635368816238	0.024850457380353\\
410.864996374674	2.66846418267278\\
353.039987091353	3.14287179614098\\
292.793272787541	0.792401333007547\\
234.817983731403	1.5208054815912\\
175.946685714034	-1.95390818422743\\
118.401129821319	-3.27182139571346\\
58.8617679005774	-8.61142169074424\\
3.98226899981637	-13.3168371394104\\
};

\addplot [color=green, dashdotted, line width=2.0pt, mark size=3.0pt, mark=asterisk, mark options={solid, green}]
  table[row sep=crcr]{%
1100.61268644082	-53.6270741724545\\
1045.74827354676	-65.0184932146331\\
989.355600393479	-49.4058953654137\\
934.184171809047	-39.6086567114592\\
876.484384885178	-30.0407231066185\\
818.499256320007	-21.0312567921742\\
760.270582863348	-9.17567704497969\\
701.028643189567	-1.07855970751075\\
644.189278385873	0.660471145546808\\
586.849250473486	-1.74999692540291\\
523.332584076725	3.79726875482083\\
465.784181789191	5.67453878409196\\
406.560890125784	9.37377037742476\\
349.497957518832	9.66315355415304\\
289.795891548648	6.65546521556917\\
233.107811747205	5.31605158403045\\
175.234366190172	-0.91156699737121\\
117.965311437874	3.63474217877797\\
58.0848046927215	8.89127396066988\\
2.90045431561794	13.7227708974785\\
};


\addplot [color=blue, dashed, line width=2.0pt, mark size=3.0pt, mark=+, mark options={solid, blue}]
  table[row sep=crcr]{%
1119.68242360017	-31.8277351851827\\
1061.34894418788	-22.4165595876241\\
1001.48711170571	-14.7891088019476\\
943.538835711023	-10.4515874099986\\
883.775312675044	-6.91804751752079\\
824.212984420631	-4.66230695200375\\
762.394224822435	-1.95477087186964\\
703.525012523727	-1.11983842692052\\
646.301917705541	-1.80256284462007\\
588.175998639937	-2.05089324547453\\
528.854914823321	1.17940528188069\\
470.18394822357	-0.160529446484134\\
410.700066071347	1.87437501461591\\
353.677753410369	1.11953117858297\\
294.238170172024	-0.833584082736706\\
235.486750873276	0.467052595171253\\
175.94783611388	-2.25829253562688\\
118.293497595017	-2.3494340631778\\
58.6948984238163	-6.12515046917792\\
3.74402714142389	-9.41605183974661\\
};

\end{axis}

\begin{axis}[%
width=1.259\fheight,
height=0.743\fheight,
at={(-0.164\fheight,-0.097\fheight)},
scale only axis,
xmin=0,
xmax=1,
ymin=0,
ymax=1,
axis line style={draw=none},
ticks=none,
axis x line*=bottom,
axis y line*=left
]
\end{axis}
\end{tikzpicture}%
        \caption{$t=1$~\si{s}}
        \label{fig:center_test_1}
    \end{subfigure} \\ [-4ex]
        \begin{subfigure}{\linewidth}
         \setlength\fheight{6 cm}
         \setlength\fwidth{\textwidth}
%
%
\definecolor{mycolor1}{rgb}{1.00000,0.00000,1.00000}%
\begin{tikzpicture}

\begin{axis}[%
width=0.976\fheight,
height=0.49\fheight,
at={(0\fheight,0\fheight)},
scale only axis,
xmin=0,
xmax=1200,
xlabel style={font=\color{white!15!black}},
xlabel={$x-$coordinate [mm]},
ymin=-120,
ymax=40,
ylabel style={font=\color{white!15!black}},
ylabel={$z-$coordinate [mm]},
axis background/.style={fill=white},
xmajorgrids,
ymajorgrids,
legend style={at={(0.174,0.184)}, anchor=south west, legend cell align=left, align=left, draw=white!15!black}
]
\addplot [color=black, line width=2.0pt, mark size=3.0pt, mark=triangle, mark options={solid, black}]
  table[row sep=crcr]{%
1102.68378930959	-4.76694459985129\\
1052.89611818097	-27.0541689220638\\
994.290947738339	-26.3457706743889\\
937.199103711481	-22.8244823465095\\
878.374925257144	-16.6443607184187\\
819.678013511813	-10.3612013162178\\
763.904682507428	-3.94900442815242\\
704.508667396595	-2.12161395878366\\
646.564046718637	-2.36563117945275\\
587.852617510001	-2.4118723387503\\
528.28963215057	1.79372799908583\\
467.882547053702	3.14324540888151\\
406.624305258093	4.72794384274607\\
348.907837154864	1.63177907178124\\
289.813778239911	0.336146583353184\\
233.431512206696	2.18502579291362\\
175.641071022283	-1.93273965081039\\
118.400211946485	-3.6943765360968\\
58.8611891445476	-9.97606306848365\\
3.99132855493417	-15.5374580760156\\
};

\addplot [color=red, dotted, line width=2.0pt, mark size=3.0pt, mark=o, mark options={solid, red}]
  table[row sep=crcr]{%
1120.85682076436	28.6611931188995\\
1056.65212227412	24.9140414026767\\
995.04257450708	16.4239545929347\\
936.045582894558	9.69050695378655\\
875.57898087947	3.71182053923303\\
814.688225296388	-2.20267048286701\\
753.115245794289	-5.34304907860928\\
696.779346582672	-0.820390632962699\\
642.709664369904	-0.405266003348288\\
587.043795409146	-2.11266854924338\\
524.829354898012	1.23197492584768\\
470.75928403173	0.608018499835225\\
411.402496903504	3.05746366484937\\
353.202279745203	2.53588708808684\\
293.42757378406	1.29365396107391\\
234.694467048904	2.34549756620049\\
175.562339034353	-1.77797593046876\\
118.378328763976	-3.49027637873064\\
58.9642126339102	-9.3130999532475\\
4.18754723927236	-14.4715599104779\\
};

\addplot [color=green, dashdotted, line width=2.0pt, mark size=3.0pt, mark=asterisk, mark options={solid, green}]
  table[row sep=crcr]{%
1098.25386177009	-80.8113369476844\\
1046.53596395012	-80.6320777599781\\
989.926999232987	-62.7715453802393\\
934.416850474398	-49.3961117702356\\
876.609828110636	-35.95981388521\\
818.85966987439	-23.3174355665533\\
762.258716129687	-8.76545082672669\\
702.569400756815	-2.03245811441684\\
645.12817644138	-0.489001899674804\\
587.225903148991	-2.22369146796632\\
524.732839601665	-0.563236266192007\\
465.461705653772	-1.52321817572374\\
405.566421987547	1.01619693516091\\
348.609522294399	1.19088899892881\\
288.238458835764	1.08387342008382\\
232.71091570557	2.38892970590587\\
175.415569741911	-1.82495622642818\\
118.517029166078	-4.04331035271366\\
59.0723178734577	-10.8286466941336\\
4.26524492709927	-16.861592578555\\
};


\addplot [color=blue, dashed, line width=2.0pt, mark size=3.0pt, mark=+, mark options={solid, blue}]
  table[row sep=crcr]{%
1114.10146865716	0.488775007046115\\
1058.20568904106	5.84192005153545\\
998.681708703953	7.51281213940115\\
940.93125611157	8.28817417062737\\
881.563455259966	7.97568903974184\\
822.375083136063	6.38320768672793\\
761.05085832166	5.37522190164486\\
702.393710818187	2.69018326552259\\
645.406816406965	-0.444656596561117\\
587.386636970936	-2.25980842218382\\
526.056099631146	0.0214641786042193\\
468.973463090643	-2.79044328399164\\
410.402083002144	-0.843428984973116\\
353.627811162191	-0.283870834816753\\
292.478786538547	-0.797155867740003\\
234.465773575506	1.22580844228264\\
175.550559260103	-1.91010228322784\\
118.511533222111	-2.06894777084585\\
58.9876362165921	-5.50520482265165\\
4.13032683736083	-8.51504221007053\\
};

\end{axis}

\begin{axis}[%
width=1.259\fheight,
height=0.743\fheight,
at={(-0.164\fheight,-0.097\fheight)},
scale only axis,
xmin=0,
xmax=1,
ymin=0,
ymax=1,
axis line style={draw=none},
ticks=none,
axis x line*=bottom,
axis y line*=left
]
\end{axis}
\end{tikzpicture}%
        \caption{$t=2$~\si{s}}
        \label{fig:center_test_2}
    \end{subfigure}
    \caption{Robustness to unseen inputs (centerline prediction for \textsf{Ep 3}). The LOpInf ROM and the DMDc ROM provide reasonable predictions up to $t=1$~\si{s} whereas the ERA/OKID ROM yields inaccurate solutions from the start. At $t=2$~\si{s}, all three data-driven linear ROMs fail to capture the twist in the centerline shape.}
    \label{fig:center_test}
\end{figure}
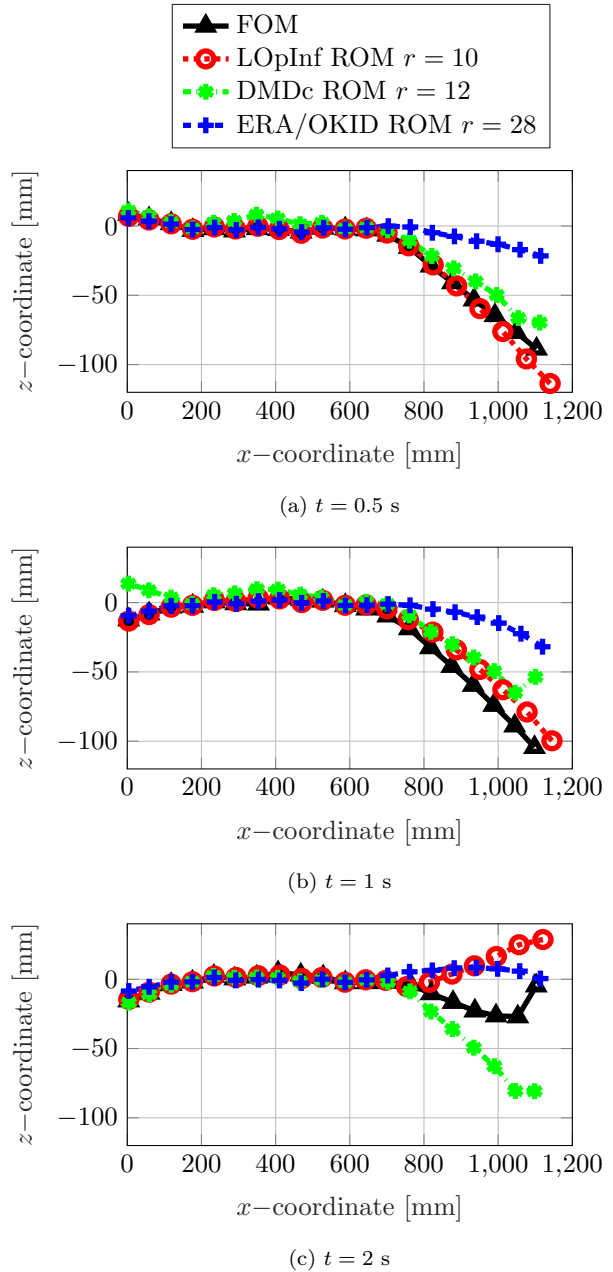

Since the ultimate aim is to employ these data-driven linear ROMs for closed-loop dynamic control of a soft robot, having a data-driven ROM that is robust to unseen inputs is desirable. Figure~\ref{fig:center_test} provides another assessment of the robustness of data-driven ROMs to unseen inputs by comparing the centerline prediction performance for \textsf{Ep 3}, a challenging test case where the soft robot exhibits strongly nonlinear behavior between $1{,}000$~\si{\mm} and  $1{,}200$~\si{\mm} from the origin in the $x-$ direction. In Figure~\ref{fig:center_test_05}, we observe that both the LOpInf ROM and the DMDc ROM capture the centerline shape accurately at $t=0.5$~\si{s} whereas the ERA/OKID ROM yields inaccurate predictions. The LOpInf ROM correctly predicts the centerline shape even at $t=1$~\si{s} in Figure~\ref{fig:center_test_1} whereas the DMDc ROM solution starts to deviate from the FOM solution $t=1$~\si{s}. 

In Figure~\ref{fig:center_test_2}, we observe that all three ROMs fail to capture the correct centerline shape at $t=2$~\si{s}. This challenging test case shows the limitations of linear data-driven ROMs in capturing strongly nonlinear behavior not observed in the training episodes.

\section{Conclusions}
\label{sec: conclusions}
We demonstrated the advantages of using Lagrangian Operator Inference, a nonintrusive structure-preserving model reduction method, for data-driven model reduction of high-dimensional soft robot models through a swimming soft robot model example with $231{,}336$ DOF. The first case study with a single training episode shows that all three methods can provide accurate prediction in the training data regime while also achieving a significant reduction in the state dimension. The second case study with multiple training episodes demonstrates the robustness to unseen inputs for both the LOpInf method and the DMDc method. These studies show that the structure-preserving LOpInf method learns generalizable data-driven linear ROMs that predict accurate solutions outside the training time interval, as well as for unseen inputs.


{\Ra{Similarly to standard data-driven linear ROMs, the LOpInf ROMs are not expected to provide accurate predictions for soft robots that exhibit significant nonlinear behavior. For future work, we plan to}} explore nonlinear model reduction techniques based on the Koopman operator theory  (\cite{bruder2019nonlinear}), {\Rb{spectral submanifold reduction (\cite{alora2023data})}}, and structure-preserving machine learning (\cite{sharma2023ml,lepri2023neural}).

\begin{ack}
H.S. and B.K. were in part financially supported by the U.S. Office of Naval Research (ONR) Grant No. N00014-22-1-2624 and the Ministry of Trade, Industry and Energy (MOTIE) and the Korea Institute for Advancement of Technology (KIAT) through the International Cooperative R\&D program (No. P0019804, Digital twin based intelligent unmanned facility inspection solutions). I.A., J.C., and M.T.T. were financially supported by the ONR Grant No. N00014-22-1-2595.
\end{ack}

\bibliography{ifacconf}             
                                                   







\appendix
\end{document}